\documentclass[11pt]{article}
\usepackage{acl2016}
\usepackage{times}
\usepackage{url}
\usepackage{latexsym}
\usepackage{amsfonts}
\usepackage{amsmath}
\usepackage{graphicx}

\aclfinalcopy %
\def\aclpaperid{142} %

\newcommand{\x}{\mathbf{x}}

\title{Exploring Prediction Uncertainty in Machine Translation \\ Quality Estimation}

\author{Daniel Beck$^\dagger$ ~~~~ Lucia Specia$^\dagger$ ~~~~ Trevor Cohn$^\ddagger$\\
  $^\dagger$Department of Computer Science\\
  University of Sheffield, United Kingdom\\
  $^\ddagger$Computing and Information Systems\\
  University of Melbourne, Australia\\
  {\tt \{debeck1,l.specia\}@sheffield.ac.uk, t.cohn@unimelb.edu.au}
}

\date{}

\begin{document}

\maketitle

\begin{abstract}
Machine Translation Quality Estimation is a notoriously difficult task, which lessens its usefulness in real-world translation environments. Such scenarios can be improved if quality predictions are accompanied by a measure of uncertainty. However, models in this task are traditionally evaluated only in terms of point estimate metrics, which do not take prediction uncertainty into account. We investigate probabilistic methods for Quality Estimation that can provide well-calibrated uncertainty estimates and evaluate them in terms of their full posterior predictive distributions. We also show how this posterior information can be useful in an asymmetric risk scenario, which aims to capture typical situations in translation workflows.%
\end{abstract}

\section{Introduction}

Quality Estimation (QE) \cite{Blatz2004,Specia2009} models aim at predicting the quality of automatically translated text segments. Traditionally, these models provide point estimates and are evaluated using metrics like Mean Absolute Error (MAE), Root-Mean-Square Error (RMSE) and Pearson's $r$ correlation coefficient. However, in practice QE models are built for use in decision making in large workflows involving Machine Translation (MT). In these settings, relying on point estimates would mean that only very accurate prediction models can be useful in practice.

A way to improve decision making based on quality predictions is to explore uncertainty estimates. Consider for example a post-editing scenario where professional translators use MT in an effort to speed-up the translation process. A QE model can be used to determine if an MT segment is good enough for post-editing or should be discarded and translated from scratch. But since QE models are not perfect they can end up allowing bad MT segments to go through for post-editing because of a prediction error. In such a scenario, having an uncertainty estimate for the prediction can provide additional information for the filtering decision. For instance, in order to ensure good user experience for the human translator and maximise translation productivity, an MT segment could be forwarded for post-editing only if a QE model assigns a high quality score with \emph{low uncertainty} (high confidence). Such a decision process is not possible with point estimates only.

Good uncertainty estimates can be acquired from well-calibrated probability distributions over the quality predictions. In QE, arguably the most successful probabilistic models are Gaussian Processes (GPs) since they considered the state-of-the-art for regression \cite{Cohn2013,Hensman2013}, especially in the low-data regimes typical for this task. We focus our analysis in this paper on GPs since other common models used in QE can only provide point estimates as predictions. Another reason why we focus on probabilistic models is because this lets us employ the ideas proposed by \newcite{Quinonero-Candela2006}, which defined new evaluation metrics that take into account probability distributions over predictions.

The remaining of this paper is organised as follows:
\begin{itemize}
\item In Section \ref{sec:probqe} we further motivate the use of GPs for uncertainty modelling in QE and revisit their underlying theory. We also propose some model extensions previously developed in the GP literature and argue they are more appropriate for the task.
\item We intrinsically evaluate our proposed models in terms of their posterior distributions on training and test data in Section \ref{sec:intrinsic}. Specifically, we show that differences in uncertainty modelling are not captured by the usual point estimate metrics commonly used for this task.
\item As an example of an application for predicitive distributions, in Section \ref{sec:asymmetric} we show how they can be useful in scenarios with asymmetric risk and how the proposed models can provide better performance in this case.
\end{itemize}
We discuss related work in Section \ref{sec:relwork} and give conclusions and avenues for future work in Section \ref{sec:conc}.

While we focus on QE as application, the methods we explore in this paper can be applied to any text regression task where modelling predictive uncertainty is useful, either in human decision making or by propagating this information for further computational processing.

\section{Probabilistic Models for QE}
\label{sec:probqe}

Traditionally, QE is treated as a regression task with hand-crafted features. Kernel methods are arguably the state-of-the-art in QE since they can easily model non-linearities in the data. Furthermore, the scalability issues that arise in kernel methods do not tend to affect QE in practice since the datasets are usually small, in the order of thousands of instances.

The most popular method for QE is Support Vector Regression (SVR), as shown in the multiple instances of the WMT QE shared tasks \cite{Callison-Burch2012,Bojar2013,Bojar2014,Bojar2015}. While SVR models can generate competitive predictions for this task, they lack a probabilistic interpretation, which makes it hard to extract uncertainty estimates using them. Bootstrapping approaches like bagging \cite{Abe1998} can be applied, but this requires setting and optimising hyperparameters like bag size and number of bootstraps. There is also no guarantee these estimates come from a well-calibrated probabilistic distribution.

Gaussian Processes (GPs) \cite{Rasmussen2006} is an alternative kernel-based framework that gives competitive results for point estimates \cite{Cohn2013,Shah2013,Beck2014b}. Unlike SVR, they explicitly model uncertainty in the data and in the predictions. This makes GPs very applicable when well-calibrated uncertainty estimates are required. Furthermore, they are very flexible in terms of modelling decisions by allowing the use of a variety of kernels and likelihoods while providing efficient ways of doing model selection. Therefore, in this work we focus on GPs for probabilistic modelling of QE. In what follows we briefly describe the GPs framework for regression.

\subsection{Gaussian Process Regression}
\label{sec:gpr}

Here we follow closely the definition of GPs given by \newcite{Rasmussen2006}. Let $\mathcal{X} = \{(\x_1, y_1),(\x_2, y_2), \dots, (\x_n, y_n) \}$ be our data, where each $\x \in \mathbb{R}^D$ is a $D$-dimensional input and $y$ is its corresponding response variable. A GP is defined as a stochastic model over the latent function $f$ that generates the data $\mathcal{X}$:
\begin{equation*}
  \label{eq:gp}
  f(\mathbf{x}) \sim \mathcal{GP} (m(\mathbf{x}), k(\mathbf{x},\mathbf{x'})),
\end{equation*}
where $m(\x)$ is the \emph{mean} function, which is usually the $0$ constant, and $k(\x,\x')$ is the kernel or \emph{covariance} function, which describes the covariance between values of $f$ at the different locations of $\x$ and $\x'$.

The prior is combined with a likelihood via Bayes' rule to obtain a posterior over the latent function:
\begin{equation*}
   \label{eq:fposterior}
   p(f|\mathcal{X}) = \frac{p(\mathbf{y}|\mathbf{X},f) p(f)}{p(\mathbf{y}|\mathbf{X})} ,
 \end{equation*}
where $\mathbf{X}$ and $\mathbf{y}$ are the training inputs and response variables, respectively. For regression, we assume that each $y_i = f(\mathbf{x_i}) + \eta$, where $\eta \sim \mathcal{N}(0,\sigma_n^2)$ is added white noise. Having a Gaussian likelihood results in a closed form solution for the posterior.

Training a GP involves the optimisation of model hyperparameters, which is done by maximising the marginal likelihood $p(\mathbf{y}|\mathbf{X})$ via gradient ascent. Predictive posteriors for unseen $\x_*$ are obtained by integrating over the latent function evaluations at $\x_*$.

GPs can be extended in many different ways by applying different kernels, likelihoods and modifying the posterior, for instance. In the next Sections, we explain in detail some sensible modelling choices in applying GPs for QE.

\subsection{Mat\`{e}rn Kernels}
\label{sec:matern-kernels}

Choosing an appropriate kernel is a crucial step in defining a GP model (and any other kernel method). A common choice is to employ the exponentiated quadratic (EQ) kernel\footnote{Also known as Radial Basis Function (RBF) kernel.}:
\begin{align*}
  \label{eq:2}
  k_{\text{EQ}}(\x, \x') &= \sigma_v \; \mathrm{exp}(-\frac{r^2}{2}) \, , \\
\mbox{where~}  r^2 &= \sum\limits_{i=1}^D\frac{(x_i - x_i')^2}{l_i^2}
\end{align*}
is the scaled distance between the two inputs,
$\sigma_v$ is a scale hyperparameter and $\mathbf{l}$ is a vector of lengthscales.
 Most kernel methods tie all lengthscale to a single value, resulting in an isotropic kernel. However, since in GPs hyperparameter optimisation can be done efficiently, it is common to employ one lengthscale per feature, a method called Automatic Relevance Determination (ARD).

The EQ kernel allows the modelling of non-linearities between the inputs and the response variables but it makes a strong assumption: it generates smooth, infinitely differentiable functions. This assumption can be too strong for noisy data. An alternative is the Mat\`{e}rn class of kernels, which relax the smoothness assumption by modelling functions which are $\nu$-times differentiable only. Common values for $\nu$ are the half-integers $3/2$ and $5/2$, resulting in the following Mat\`{e}rn kernels:
\begin{align*}
  k_{\text{M32}} &= \sigma_v (1 + \sqrt{3r^2}) \; \mathrm{exp}(-\sqrt{3r^2}) \\
  k_{\text{M52}} &= \sigma_v \left(1 + \sqrt{5r^2} + \frac{5r^2}{3}\right) \mathrm{exp}(-\sqrt{5r^2}) \, ,
\end{align*}
where we have omitted the dependence of $k_{\text{M32}}$ and $k_{\text{M52}}$ on the inputs $(\x, \x')$ for brevity.
Higher values for $\nu$ are usually not very useful since the resulting behaviour is hard to distinguish from limit case $\nu \rightarrow \infty$, which retrieves the EQ kernel \cite[Sec. 4.2]{Rasmussen2006}.

The relaxed smoothness assumptions from the Mat\`{e}rn kernels makes them promising candidates for QE datasets, which tend to be very noisy. We expect that employing them will result in a better models for this application.

\subsection{Warped Gaussian Processes}
\label{sec:wgp}

The Gaussian likelihood of standard GPs has support over the entire real number line. However, common quality scores are strictly positive values, which means that the Gaussian assumption is not ideal. A usual way to deal with this problem is model the logarithm of the response variables, since this transformation maps strictly positive values to the real line. However, there is no reason to believe this is the best possible mapping: a better idea would be to learn it from the data.

Warped GPs \cite{Snelson2004} are an extension of GPs that allows the learning of arbitrary mappings. It does that by placing a monotonic \emph{warping function} over the observations and modelling the warped values inside a standard GP. The posterior distribution is obtained by applying a change of variables:
\begin{equation*}
  p(y_*|\x_*) = \frac{f'(y_*)}{\sqrt{2\pi\sigma_*^2}} \; \mathrm{exp} \left(\frac{f(y_*) - \mu_*}{2\sigma_*}\right),
\end{equation*}
where $\mu_*$ and $\sigma_*$ are the mean and standard deviation of the latent (warped) response variable and $f$ and $f'$ are the warping function and its derivative.

Point predictions from this model depend on the loss function to be minimised. For absolute error, the median is the optimal value while for squared error it is the mean of the posterior. In standard GPs, since the posterior is Gaussian the median and mean coincide but this in general is not the case for a Warped GP posterior. The median can be easily obtained by applying the inverse warping function to the latent median:
\begin{equation*}
  y^{\mathrm{med}}_* = f^{-1}(\mu_*).
\end{equation*}
While the inverse of the warping function is usually not available in closed form, we can use its gradient to have a numerical estimate.

The mean is obtained by integrating $y^*$ over the latent density:
\begin{equation*}
  \mathbb{E}[y_*] = \int f^{-1}(z) \mathcal{N}_z(\mu_*, \sigma^2_*) dz,
\end{equation*}
where $z$ is the latent variable. This can be easily approximated using Gauss-Hermite quadrature since it is a one dimensional integral over a Gaussian density.

The warping function should be flexible enough to allow the learning of complex mappings, but it needs to be monotonic. \newcite{Snelson2004} proposes a parametric form composed of a sum of $\mathrm{tanh}$ functions, similar to a neural network layer:
\begin{equation*}
  \label{eq:warp}
  f(y) = y + \sum\limits_{i=1}^{I} a_i \; \mathrm{tanh} (b_i (y + c_i)) ,
\end{equation*}
where $I$ is the number of $\mathrm{tanh}$ terms and $\mathbf{a}, \mathbf{b}$ and $\mathbf{c}$ are treated as model hyperparameters and optimised jointly with the kernel and likelihood hyperparameters. Large values for $I$ allow more complex mappings to be learned but raise the risk of overfitting.

Warped GPs provide an easy and elegant way to model response variables with non-Gaussian behaviour within the GP framework. In our experiments we explore models employing warping functions with up to $3$ terms, which is the value recommended by \newcite{Snelson2004}.  We also report results using the $f(y) = \log(y)$ warping function.

\section{Intrinsic Uncertainty Evaluation}
\label{sec:intrinsic}

Given a set of different probabilistic QE models, we are interested in evaluating the performance of these models, while also taking their uncertainty into account, particularly to distinguish among models with seemingly same or similar performance. A straightforward way to measure the performance of a probabilistic model is to inspect its negative ($\mathrm{log}$) marginal likelihood. This measure, however, does not capture if a model overfit the training data.

We can have a better generalisation measure by calculating the likelihood on \emph{test data} instead. This was proposed in previous work and it is called Negative Log Predictive Density (NLPD) \cite{Quinonero-Candela2006}:
\begin{equation*}
  \label{eq:nlpd}
  \text{NLPD}(\mathbf{\hat{y}}, \mathbf{y}) = -\frac{1}{n} \sum\limits_{i=1}^n \mathrm{log}\; p(\hat{y}_i = y_i|\x_i).
\end{equation*}
where $\mathbf{\hat{y}}$ is a set of test predictions, $\mathbf{y}$ is the set of true labels and $n$ is the test set size. This metric has since been largely adopted by the ML community when evaluating GPs and other probabilistic models for regression (see Section \ref{sec:relwork} for some examples).

As with other error metrics, lower values are better. Intuitively, if two models produce equally incorrect predictions but they have different uncertainty estimates, NLPD will penalise the overconfident model more than the underconfident one. On the other hand, if predictions are close to the true value then NLPD will penalise the underconfident model instead.

In our first set of experiments we evaluate models proposed in Section \ref{sec:probqe} according to their negative $\mathrm{log}$ likelihood (NLL) and the NLPD on test data. We also report two point estimate metrics on test data: Mean Absolute Error (MAE), the most commonly used evaluation metric in QE, and Pearson's $r$, which has recently proposed by \newcite{Graham2015} as a more robust alternative.

\subsection{Experimental Settings}
\label{sec:exp}

Our experiments comprise datasets containing three different language pairs, where the label to predict is post-editing time:
\begin{description}
\item[English-Spanish (en-es)] This dataset was used in the WMT14 QE shared task \cite{Bojar2014}. It contains $858$ sentences translated by one MT system and post-edited by a professional translator.
\item[French-English (fr-en)] Described in \cite{Specia2011}, this dataset contains $2,525$ sentences translated by one MT system and post-edited by a professional translator. 
\item[English-German (en-de)] This dataset is part of the WMT16 QE shared task\footnote{\url{www.statmt.org/wmt16}}. It was translated by one MT system for consistency we use a subset of $2,828$ instances post-edited by a single professional translator.
\end{description}

As part of the process of creating these datasets, post-editing time was logged on an sentence basis for all datasets. Following common practice, we normalise the post-editing time by the length of the machine translated sentence to obtain post-editing {\em rates} and use these as our response variables.

Technically our approach could be used with any other numeric quality labels from the literature, including the commonly used Human Translation Error Rate (HTER) \cite{Snover2006}. Our decision to focus on post-editing time  was based on the fact that time is a more complete measure of post-editing effort, capturing not only technical effort like HTER, but also cognitive effort \cite{Koponen2012}. Additionally, time is more directly applicable in real translation environments -- where uncertainty estimates could be useful, as it relates directly to productivity measures.

For model building, we use a standard set of $17$ features from the QuEst framework \cite{Specia2015}. These features are used in the strong baseline models provided by the WMT QE shared tasks. While the best performing systems in the shared tasks use larger feature sets, these are mostly resource-intensive and language-dependent, and therefore not equally applicable to all our language pairs.
Moreover, our goal is to compare probabilistic QE models through the predictive uncertainty perspective, rather than improving the state-of-the-art in terms of point predictions.
We perform $10$-fold cross validation instead of using a single train/test splits and report averaged metric scores. 

The model hyperparameters were optimised by maximising the likelihood on the training data. We perform a two-pass procedure similar to that in \cite{Cohn2013}: first we employ an isotropic kernel and optimise all hyperparameters using $10$ random restarts; then we move to an ARD equivalent kernel and perform a final optimisation step to fine tune feature {\em lengthscales}. Point predictions were fixed as the median of the distribution.

\subsection{Results and Discussion}
\label{sec:iresults}

Table \ref{tab:intrinsic} shows the results obtained for all datasets.
The first two columns shows an interesting finding in terms of model learning: using a warping function drastically decreases both NLL and NLPD. The main reason behind this is that standard GPs distribute probability mass over negative values, while the warped models do not. For the {\bf fr-en} and {\bf en-de} datasets, NLL and NLPD follow similar trends. This means that we can trust NLL as a measure of uncertainty for these datasets. However, this is not observed in the {\bf en-es} dataset. Since this dataset is considerably smaller than the others, we believe this is evidence of overfitting, thus showing that NLL is not a reliable metric for small datasets.

\begin{table}[ht!]
  \centering
  \begin{small}
  \begin{tabular}{|l|c|c|c|c|}
    \hline
    \multicolumn{5}{|l|}{\bf English-Spanish - 858 instances} \\
    \hline
    \hline
    & NLL & NLPD & MAE & $r$\\
    \hline
    \hline
    EQ & 1244.03 & 1.632 & 0.828 & 0.362 \\
    Mat32 & 1237.48 & 1.649 & 0.862 & 0.330 \\
    Mat52 & 1240.76 & 1.637 & 0.853 & 0.340 \\
    \hline
    \hline
    log EQ & 986.14 & 1.277 & 0.798 & 0.368 \\
    log Mat32 & 982.71 & 1.271 & 0.793 & 0.380 \\
    log Mat52 & 982.31 & 1.272 & 0.794 & 0.376 \\
    \hline
    tanh1 EQ & 992.19 & 1.274 & 0.790 & 0.375 \\
    tanh1 Mat32 & 991.39 & 1.272 & 0.790 & 0.379 \\
    tanh1 Mat52 & 992.20 & 1.274 & 0.791 & 0.376 \\
    \hline
    tanh2 EQ & 982.43 & 1.275 & 0.792 & 0.376 \\
    tanh2 Mat32 & 982.40 & 1.281 & 0.791 & 0.382 \\
    tanh2 Mat52 & 981.86 & 1.282 & 0.792 & 0.278 \\
    \hline
    tanh3 EQ & 980.50 & 1.282 & 0.791 & 0.380 \\
    tanh3 Mat32 & 981.20 & 1.282 & 0.791 & 0.380 \\
    tanh3 Mat52 & 980.70 & 1.275 & 0.790 & 0.385 \\
    \hline
    \multicolumn{5}{l}{}\\
    \hline
    \multicolumn{5}{|l|}{\bf French-English - 2525 instances} \\
    \hline
    \hline
    & NLL & NLPD & MAE & $r$\\
    \hline
    \hline
    EQ & 2334.17 & 1.039 & 0.491 & 0.322 \\
    Mat32 & 2335.81 & 1.040 & 0.491 & 0.320 \\
    Mat52 & 2344.86 & 1.037 & 0.490 & 0.320 \\
    \hline
    \hline
    log EQ & 1935.71 & 0.855 & 0.493 & 0.314 \\
    log Mat32 & 1949.02 & 0.857 & 0.493 & 0.310 \\
    log Mat52 & 1937.31 & 0.855 & 0.493 & 0.313 \\
    \hline
    tanh1 EQ & 1884.82 & 0.840 & 0.482 & 0.322 \\
    tanh1 Mat32 & 1890.34 & 0.840 & 0.482 & 0.317 \\
    tanh1 Mat52 & 1887.41 & 0.834 & 0.482 & 0.320 \\
    \hline
    tanh2 EQ & 1762.33 & 0.775 & 0.483 & 0.323 \\
    tanh2 Mat32 & 1717.62 & 0.754 & 0.483 & 0.313 \\
    tanh2 Mat52 & 1748.62 & 0.768 & 0.486 & 0.306 \\
    \hline
    tanh3 EQ & 1814.99 & 0.803 & 0.484 & 0.314 \\
    tanh3 Mat32 & 1723.89 & 0.760 & 0.486 & 0.302 \\
    tanh3 Mat52 & 1706.28 & 0.751 & 0.482 & 0.320 \\
    \hline
    \multicolumn{5}{l}{}\\
    \hline
    \multicolumn{5}{|l|}{\bf English-German - 2828 instances} \\
    \hline
    \hline
    & NLL & NLPD & MAE & $r$\\
    \hline
    \hline
    EQ & 4852.80 & 1.865 & 1.103 & 0.359 \\
    Mat32 & 4850.27 & 1.861 & 1.098 & 0.369 \\
    Mat52 & 4850.33 & 1.861 & 1.098 & 0.369 \\
    \hline
    \hline
    log EQ & 4053.43 & 1.581 & 1.063 & 0.360 \\
    log Mat32 & 4054.51 & 1.580 & 1.063 & 0.363 \\
    log Mat52 & 4054.39 & 1.581 & 1.064 & 0.363 \\
    \hline
    tanh1 EQ & 4116.86 & 1.597 & 1.068 & 0.343 \\
    tanh1 Mat32 & 4113.74 & 1.593 & 1.064 & 0.351 \\
    tanh1 Mat52 & 4112.91 & 1.595 & 1.068 & 0.349 \\
    \hline
    tanh2 EQ & 4032.70 & 1.570 & 1.060 & 0.359 \\
    tanh2 Mat32 & 4031.42 & 1.570 & 1.060 & 0.362 \\
    tanh2 Mat52 & 4032.06 & 1.570 & 1.060 & 0.361 \\
    \hline
    tanh3 EQ & 4023.72 & 1.569 & 1.062 & 0.359 \\
    tanh3 Mat32 & 4024.64 & 1.567 & 1.058 & 0.364 \\
    tanh3 Mat52 & 4026.07 & 1.566 & 1.059 & 0.365 \\
    \hline

  \end{tabular}
  \end{small}
  \caption{Intrinsic evaluation results. The first three rows in each table correspond to standard GP models, while the remaining rows are Warped GP models with different warping functions. The number after the $\mathrm{tanh}$ models shows the number of terms in the warping function (see Equation \ref{eq:warp}). All $r$ scores have $p < 0.05$.}
  \label{tab:intrinsic}
\end{table}

In terms of different warping functions, using the parametric $\mathrm{tanh}$ function with $3$ terms performs better than the $\mathrm{log}$ for the {\bf fr-en} and {\bf en-de} datasets. This is not the case of the {\bf en-es} dataset, where the $\mathrm{log}$ function tends to perform better. We believe that this is again due to the smaller dataset size. The gains from using a Mat\`{e}rn kernel over EQ are less conclusive. While they tend to perform better for {\bf fr-en}, there does not seem to be any difference in the other datasets. %
Different kernels can be more appropriate depending on the language pair, but more experiments are needed to verify this, which we leave for future work.

The differences in uncertainty modelling are by and large not captured by the point estimate metrics. While MAE does show gains from standard to Warped GPs, it does not reflect the difference found between warping functions for {\bf fr-en}. Pearson's $r$ is also quite inconclusive in this sense, except for some observed gains for {\bf en-es}. This shows that NLPD indeed should be preferred as a evaluation metric when proper prediction uncertainty estimates are required by a QE model.

\subsection{Qualitative Analysis}
\label{sec:analysis}

\begin{figure}[ht!]
  \centering
  \includegraphics[scale=0.46]{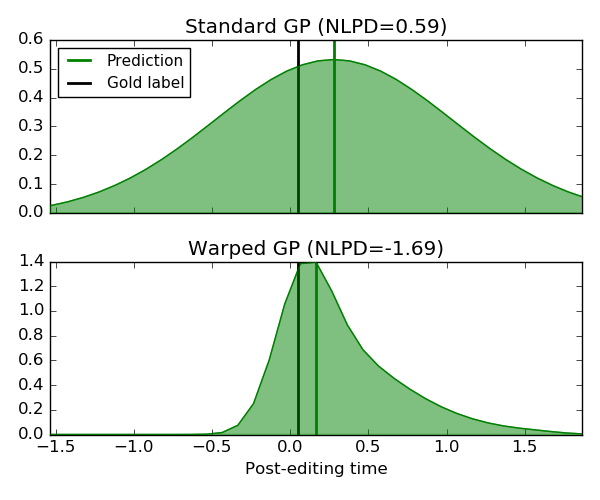}

  \includegraphics[scale=0.46]{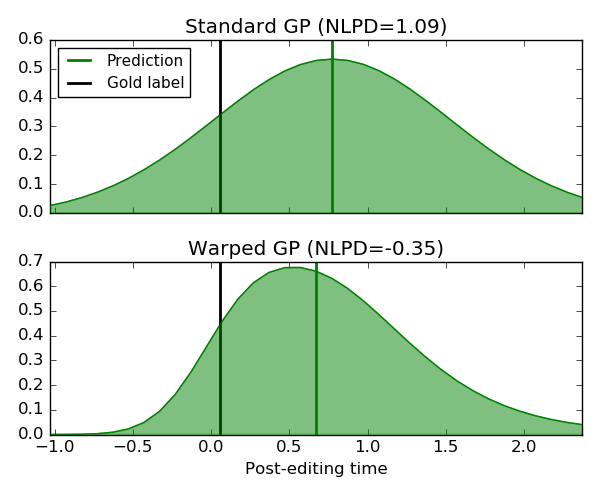}
  \caption{Predictive distributions for two {\bf fr-en} instances under a Standard GP and a Warped GP. The top two plots correspond to a prediction with low absolute error, while the bottom two plots show the behaviour when the absolute error is high.}%
  \label{fig:example1}
\end{figure}

To obtain more insights about the performance in uncertainty modelling we inspected the predictive distributions for two sentence pairs in the {\bf fr-en} dataset. We show the distributions for a standard GP and a Warped GP with a $\mathrm{tanh3}$ function in Figure \ref{fig:example1}. In the first case, where both models give accurate predictions, we see that the Warped GP distribution is peaked around the predicted value, as it should be. It also gives more probability mass to positive values, showing that the model is able to learn that the label is non-negative. In the second case we analyse the distributions when both models make inaccurate predictions. We can see that the Warped GP is able to give a broader distribution in this case, while still keeping most of the mass outside the negative range.

We also report above each plot in Figure \ref{fig:example1} the NLPD for each prediction. Comparing only the Warped GP predictions, we can see that their values reflect the fact that we prefer sharp distributions when predictions are accurate and broader ones when predictions are not accurate. However, it is interesting to see that the metric also penalises predictions when their distributions are too broad, as it is the case with the standard GPs since they can not discriminate between positive and negative values as well as the Warped GPs.

\begin{figure}[ht!]
   \centering
   \includegraphics[scale=0.42]{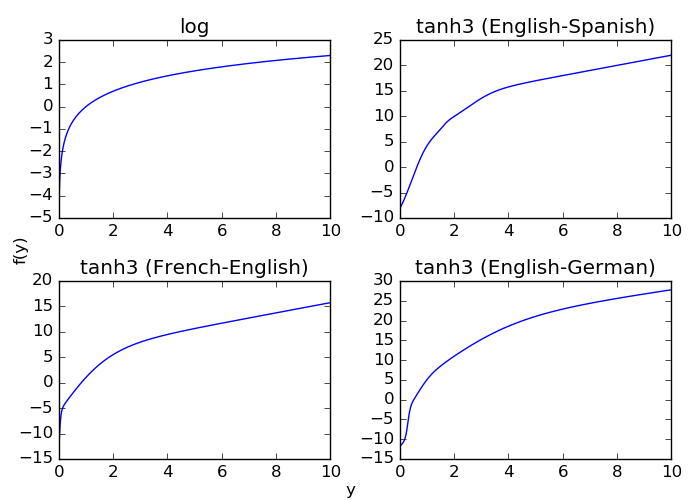}
   \caption{Warping function instances from the three datasets. The vertical axis correspond to the latent warped values. The horizontal axis show the observed response variables, which are always positive in our case since they are post-editing times.}
   \label{fig:warps}
 \end{figure}

Inspecting the resulting warping functions can bring additional modelling insights. In Figure \ref{fig:warps} we show  instances of $\mathrm{tanh3}$ warping functions learned from the three datasets and compare them with the $\mathrm{log}$ warping function. %
We can see that the parametric $\mathrm{tanh3}$ model is able to learn non-trivial mappings. For instance, in the {\bf en-es} case %
the learned function is roughly logarithmic in the low scales but it switches to a linear mapping after $y = 4$. Notice also the difference in the scales, which means that the optimal model uses a latent Gaussian with a larger variance.%

\section{Asymmetric Risk Scenarios}
\label{sec:asymmetric}

Evaluation metrics for QE, including those used in the WMT QE shared tasks, are assumed to be symmetric, i.e., they penalise over and underestimates equally. This assumption is however too simplistic for many possible applications of QE. For example:
\begin{itemize}
\item In a {\em post-editing} scenario, a project manager may have translators with limited expertise in post-editing. In this case, automatic translations should not be provided to the translator unless they are highly likely to have very good quality. This can be enforced this by increasing the penalisation weight for underestimates. We call this the \emph{pessimistic} scenario.
\item In a {\em gisting} scenario, a company wants to automatically translate their product reviews so that they can be published in a foreign language without human intervention. 
The company would prefer to publish only the reviews translated well enough, but having more reviews published will increase the chances of selling products. In this case, having better recall is more important and thus only  reviews with very poor translation quality should be discarded. We can accomplish this by heavier penalisation on overestimates, a scenario we call \emph{optimistic}.
\end{itemize}

In this Section we show how these scenarios can be addressed by well-calibrated predictive distributions and by employing {\em asymmetric} loss functions. An example of such a function is the asymmetric linear (henceforth, AL) loss, which is a generalisation of the absolute error:
\begin{equation*}
  \label{eq:asymmae}
  L(\hat{y}, y) =  \begin{cases}
    w(\hat{y} - y) &\text{if } \hat{y} > y\\
    y - \hat{y} &\text{if } \hat{y} \le y ,
  \end{cases}
\end{equation*}
where $w > 0$ is the weight given to overestimates. If $w > 1$ we have the pessimistic scenario, and the optimistic one can be obtained using $0 < w < 1$. For $w = 1$ we retrieve the original absolute error loss.

\begin{figure}[ht!]
  \centering
  \includegraphics[scale=0.5]{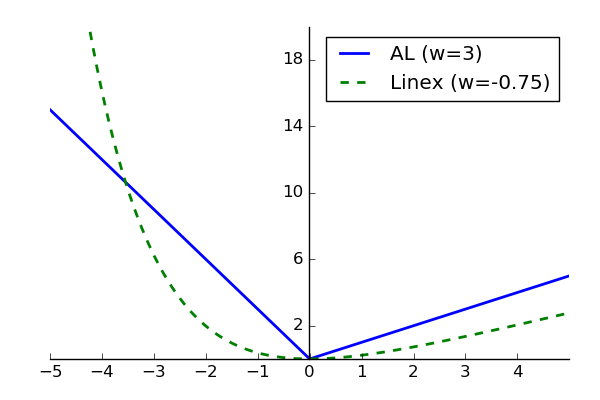}
  \caption{Asymmetric losses. These curves correspond to the pessimistic scenario since they impose larger penalties when the prediction is lower than the true label. In the optimistic scenario the curves would be reflected with respect to the vertical axis.}
  \label{fig:losses}
\end{figure}

Another asymmetric loss is the linear exponential or {\em linex} loss \cite{Zellner1986}:
\begin{equation*}
  \label{eq:linex}
  L(\hat{y}, y) = \mathrm{exp}[w(\hat{y} - y)] - (\hat{y} - y) - 1
\end{equation*}
where $w \in \mathbb{R}$ is the weight. This loss attempts to keep a linear penalty in lesser risk regions, while imposing an exponential penalty in the higher risk ones. Negative values for $w$ will result in a pessimistic setting, while positive values will result in the optimistic one. For $w = 0$, the loss approximates a squared error loss. Usual values for $w$ tend to be close to $1$ or $-1$ since for higher weights the loss can quickly reach very large scores. Both losses are shown on Figure \ref{fig:losses}.

\subsection{Bayes Risk for Asymmetric Losses}
\label{sec:risk}

The losses introduced above can be incorporated directly into learning algorithms to obtain models for a given scenario. In the context of the AL loss this is called {\em quantile regression} \cite{Koenker2005}, since optimal estimators for this loss are posterior quantiles. However, in a production environment the loss can change over time. For instance, in the gisting scenario discussed above the parameter $w$ could be changed based on feedback from indicators of sales revenue or user experience. If the loss is attached to the underlying learning algorithms, a change in $w$ would require full model retraining, which can be costly.

Instead of retraining the model every time there is a different loss, we can train a single probabilistic model and derive Bayes risk estimators for the loss we are interested in. This allows estimates to be obtained without having to retrain models when the loss changes. Additionally, this allows different losses/scenarios to be employed at the same time using the same model.%

Minimum Bayes risk estimators for asymmetric losses were proposed by \newcite{Christoffersen1997} and we follow their derivations in our experiments. The best estimator for the AL loss is equivalent to the $\frac{w}{w + 1}$ quantile of the predictive distribution. Note that we retrieve the median when $w = 1$, as expected. The best estimator for the linex loss can be easily derived and results in:
\begin{equation*}
  \hat{y} = \mu_y - \frac{w \sigma^2_y}{2}
\end{equation*}
where $\mu_y$ and $\sigma^2_y$ are the mean and the variance of the predictive posterior.

\subsection{Experimental Settings}
\label{sec:exp2}

Here we assess the models and datasets used in Section \ref{sec:exp} in terms of their performance in the asymmetric setting. Following the explanation in the previous Section, we do not perform any retraining: we collect the predictions obtained using the 10-fold cross-validation protocol and apply different Bayes estimators corresponding to the asymmetric losses. Evaluation is performed using the same loss employed in the estimator (for instance, when using the linex estimator with $w = 0.75$ we report the results using the linex loss with same $w$) and averaged over the 10 folds.

To simulate both pessimistic and optimistic scenarios, we use $w \in \{ 3, 1/3 \}$ for the AL loss and $w \in \{-0.75, 0.75\}$ for the linex loss. The only exception is the {\bf en-de} dataset, where we report results for $w \in {-0.25, 0.75}$ for linex\footnote{Using $w = -0.75$ in this case resulted in loss values on the order of $10^7$. In fact, as it will be discussed in the next Section, the results for the linex loss in the pessimistic scenario were inconclusive. However, we report results using a higher $w$ in this case for completeness and to clarify the inconclusive trends we found.}.
We also report results only for models using the Mat\`{e}rn52 kernel. While we did experiment with different kernels and weighting schemes\footnote{We also tried $w \in \{1/9, 1/7, 1/5, 5, 7, 9\}$ for the AL loss and $w \in \{-0.5, -0.25, 0.25, 0.5\}$ for the linex loss.} our findings showed similar trends so we omit them for the sake of clarity.

\subsection{Results and Discussion}
\label{sec:results2}

Results are shown on Table \ref{tab:asymm}. In the optimistic scenario the $\mathrm{tanh}$-based warped GP models give consistently better results than standard GPs. The $\mathrm{log}$-based models also gives good results for AL but for linex the results are mixed except for en-es. This is probably again related to the larger sizes of the fr-en and en-de datasets, which allows the $\mathrm{tanh}$-based models to learn richer representations.

\begin{table}[ht!]
  \centering
  \begin{tabular}{|l|c|c|c|c|}
    \hline
    \multicolumn{5}{|l|}{\bf English-Spanish} \\
    \hline
    & \multicolumn{2}{c|}{\bf Optimistic} & \multicolumn{2}{c|}{\bf Pessimistic} \\
    \hline
    & AL & Linex & AL & Linex\\
    \hline
    \hline
    Std GP & 1.187 & 0.447 & 1.633 & 3.009 \\
    \hline
    log & 1.060 & 0.299 & 1.534 & 3.327 \\
    tanh1 & 1.050 & 0.300 & 1.528 & 3.251 \\
    tanh2 & 1.054 & 0.300 & 1.543 & 3.335 \\
    tanh3 & 1.053 & 0.299 & 1.538 & 3.322 \\
    \hline
    \multicolumn{5}{l}{}\\
    \hline
    \multicolumn{5}{|l|}{\bf French-English} \\
    \hline
    \hline
    & \multicolumn{2}{c|}{\bf Optimistic} & \multicolumn{2}{c|}{\bf Pessimistic} \\
    \hline
    & AL & Linex & AL & Linex\\
    \hline
    \hline
    Std GP & 0.677 & 0.127 & 0.901 & 0.337 \\
    \hline
    log & 0.675 & 0.161 & 0.914 & 0.492 \\
    tanh1 & 0.677 & 0.124 & 0.901 & 0.341 \\
    tanh2 & 0.671 & 0.121 & 0.894 & 0.347 \\
    tanh3 & 0.666 & 0.120 & 0.886 & 0.349 \\
    \hline
    \multicolumn{5}{l}{}\\
    \hline
    \multicolumn{5}{|l|}{\bf English-German} \\
    \hline
    \hline
    & \multicolumn{2}{c|}{\bf Optimistic} & \multicolumn{2}{c|}{\bf Pessimistic} \\
    \hline
    & AL & Linex & AL & Linex\\
    \hline
    \hline
    Std GP & 1.528 & 0.610 & 2.120 & 0.217 \\
    \hline
    log & 1.457 & 0.537 & 2.049 & 0.222 \\
    tanh1 & 1.459 & 0.503 & 2.064 & 0.220 \\
    tanh2 & 1.455 & 0.504 & 2.045 & 0.220 \\
    tanh3 & 1.456 & 0.497 & 2.042 & 0.219 \\
    \hline

  \end{tabular}
  \caption{Asymmetric loss experiments results. The first line in each table corresponds to a standard GP while the others are Warped GPs with different warping functions. All models use the Mat\`{e}rn52 kernel. The optimistic setting corresponds to $w = 1/3$ for AL and $w = 0.75$ for linex. The pessimistic setting uses $w = 3$ for AL and $w = -0.75$ for linex, except for English-German, where $w = -0.25$.  }
  \label{tab:asymm}
\end{table}

The pessimistic scenario shows interesting trends. While the results for AL follow a similar pattern when compared to the optimistic setting, the results for linex are consistently worse than the standard GP baseline. A key difference between AL and linex is that the latter depends on the variance of the predictive distribution. Since the warped models tend to have less variance, we believe the estimator is not being ``pushed'' towards the positive tails as much as in the standard GPs. This turns the resulting predictions not conservative enough (i.e. the post-editing time predictions are lower) and this is heavily (exponentially) penalised by the loss. This might be a case where a standard GP is preferred but can also indicate that this loss is biased towards models with high variance, even if it does that by assigning probability mass to nonsensical values (like negative time). We leave further investigation of this phenomenon for future work.

\section{Related Work}
\label{sec:relwork}

Quality Estimation is generally framed as text regression task, similarly to many other applications such as movie revenue forecasting based on reviews \cite{Joshi2010,Bitvai2015} and detection of emotion strength in news headlines \cite{Strapparava2008,Beck2014a} and song lyrics \cite{Mihalcea2012}. In general, these applications are evaluated in terms of their point estimate predictions, arguably because not all of them employ probabilistic models.

The NLPD is common and established metric used in the GP literature to evaluate new approaches. Examples include the original work on Warped GPs \cite{Snelson2004}, but also others like \newcite{Lazaro-Gredilla2012} and \newcite{Chalupka2013}. It has also been used to evaluate recent work on uncertainty propagation methods for neural networks \cite{Hernandez-Lobato2015}.

Asymmetric loss functions are common in the econometrics literature and were studied by \newcite{Zellner1986} and \newcite{Koenker2005}, among others. Besides the AL and the linex, another well studied loss is the asymmetric quadratic, which in turn relates to the concept of {\em expectiles} \cite{Newey1987}. This loss generalises the commonly used squared error loss. In terms of applications, \newcite{Cain1995} gives an example in real estate assessment, where the consequences of under- and over-assessment are usually different depending on the specific scenario. An engineering example is given by \newcite{Zellner1986} in the context of dam construction, where an underestimate of peak water level is much more serious than an overestimate. Such real-world applications guided many developments in this field: we believe that translation and other language processing scenarios which rely on NLP technologies can heavily benefit from these advancements.

\section{Conclusions}
\label{sec:conc}

This work explored new probabilistic models for machine translation QE that allow better uncertainty estimates. We proposed the use of NLPD, which can capture information on the whole predictive distribution, unlike usual point estimate-based metrics. By assessing models using NLPD we can make better informed decisions about which model to employ for different settings. Furthermore, we showed how information in the predictive distribution can be used in asymmetric loss scenarios and how the proposed models can be beneficial in these settings.

Uncertainty estimates can be useful in many other settings beyond the ones explored in this work. Active Learning can benefit from variance information in their query methods and it has shown to be useful for QE \cite{Beck2013}. Exploratory analysis is another avenue for future work, where error bars can provide further insights about the task, as shown in recent work \cite{Nguyen2015}. This kind of analysis can be useful for tracking post-editor behaviour and assessing cost estimates for translation projects, for instance.

Our main goal in this paper was to raise awareness about how different modelling aspects should be taken into account when building QE models. Decision making can be risky using simple point estimates and we believe that uncertainty information can be beneficial in such scenarios by providing more informed solutions. These ideas are not restricted to QE and we hope to see similar studies in other natural language applications in the future.%

\section*{Acknowledgements}

Daniel Beck was supported by funding from CNPq/Brazil (No. 237999/2012-9). Lucia Specia was supported by the QT21 project (H2020 No. 645452). Trevor Cohn is the recipient of an Australian Research Council Future Fellowship (project number FT130101105). The authors would like to thank James Hensman for his advice on Warped GPs and the three anonymous reviewers for their comments.

\bibliography{books,my,qe,gp,reg,sa,al,tools}

\begin{thebibliography}{}

\bibitem[\protect\citename{Abe and Mamitsuka}1998]{Abe1998}
Naoki Abe and Hiroshi Mamitsuka.
\newblock 1998.
\newblock {Query learning strategies using boosting and bagging}.
\newblock In {\em Proceedings of the Fifteenth International Conference on
  Machine Learning}, pages 1--9.

\bibitem[\protect\citename{Beck \bgroup et al.\egroup }2013]{Beck2013}
Daniel Beck, Lucia Specia, and Trevor Cohn.
\newblock 2013.
\newblock {Reducing Annotation Effort for Quality Estimation via Active
  Learning}.
\newblock In {\em Proceedings of ACL}.

\bibitem[\protect\citename{Beck \bgroup et al.\egroup }2014a]{Beck2014a}
Daniel Beck, Trevor Cohn, and Lucia Specia.
\newblock 2014a.
\newblock {Joint Emotion Analysis via Multi-task Gaussian Processes}.
\newblock In {\em Proceedings of EMNLP}, pages 1798--1803.

\bibitem[\protect\citename{Beck \bgroup et al.\egroup }2014b]{Beck2014b}
Daniel Beck, Kashif Shah, and Lucia Specia.
\newblock 2014b.
\newblock {SHEF-Lite 2.0 : Sparse Multi-task Gaussian Processes for Translation
  Quality Estimation}.
\newblock In {\em Proceedings of WMT14}, pages 307--312.

\bibitem[\protect\citename{Bitvai and Cohn}2015]{Bitvai2015}
Zsolt Bitvai and Trevor Cohn.
\newblock 2015.
\newblock {Non-Linear Text Regression with a Deep Convolutional Neural
  Network}.
\newblock In {\em Proceedings of ACL}.

\bibitem[\protect\citename{Blatz \bgroup et al.\egroup }2004]{Blatz2004}
John Blatz, Erin Fitzgerald, and George Foster.
\newblock 2004.
\newblock {Confidence estimation for machine translation}.
\newblock In {\em Proceedings of the 20th Conference on Computational
  Linguistics}, pages 315--321.

\bibitem[\protect\citename{Bojar \bgroup et al.\egroup }2013]{Bojar2013}
Ondřej Bojar, Christian Buck, Chris Callison-Burch, Christian Federmann, Barry
  Haddow, Philipp Koehn, Christof Monz, Matt Post, Radu Soricut, and Lucia
  Specia.
\newblock 2013.
\newblock {Findings of the 2013 Workshop on Statistical Machine Translation}.
\newblock In {\em Proceedings of WMT13}, pages 1--44.

\bibitem[\protect\citename{Bojar \bgroup et al.\egroup }2014]{Bojar2014}
Ondřej Bojar, Christian Buck, Christian Federmann, Barry Haddow, Philipp
  Koehn, Johannes Leveling, Christof Monz, Pavel Pecina, Matt Post, Herve
  Saint-amand, Radu Soricut, Lucia Specia, and Ale{\v{s}} Tamchyna.
\newblock 2014.
\newblock {Findings of the 2014 Workshop on Statistical Machine Translation}.
\newblock In {\em Proceedings of WMT14}, pages 12--58.

\bibitem[\protect\citename{Bojar \bgroup et al.\egroup }2015]{Bojar2015}
Ondřej Bojar, Rajen Chatterjee, Christian Federmann, Barry Haddow, Matthias
  Huck, Chris Hokamp, Philipp Koehn, Varvara Logacheva, Christof Monz, Matteo
  Negri, Matt Post, Carolina Scarton, Lucia Specia, and Marco Turchi.
\newblock 2015.
\newblock {Findings of the 2015 Workshop on Statistical Machine Translation}.
\newblock In {\em Proceedings of WMT15}, pages 22--64.

\bibitem[\protect\citename{Cain and Janssen}1995]{Cain1995}
Michael Cain and Christian Janssen.
\newblock 1995.
\newblock {Real Estate Price Prediction under Asymmetric Loss}.
\newblock {\em Annals of the Institute of Statististical Mathematics},
  47(3):401--414.

\bibitem[\protect\citename{Callison-burch \bgroup et al.\egroup
  }2012]{Callison-Burch2012}
Chris Callison-burch, Philipp Koehn, Christof Monz, Matt Post, Radu Soricut,
  and Lucia Specia.
\newblock 2012.
\newblock {Findings of the 2012 Workshop on Statistical Machine Translation}.
\newblock In {\em Proceedings of WMT12}.

\bibitem[\protect\citename{Chalupka \bgroup et al.\egroup }2013]{Chalupka2013}
Krzysztof Chalupka, Christopher K.~I. Williams, and Iain Murray.
\newblock 2013.
\newblock {A Framework for Evaluating Approximation Methods for Gaussian
  Process Regression}.
\newblock {\em Journal of Machine Learning Research}, 14:333--350.

\bibitem[\protect\citename{Christoffersen and Diebold}1997]{Christoffersen1997}
Peter~F. Christoffersen and Francis~X. Diebold.
\newblock 1997.
\newblock {Optimal Prediction Under Asymmetric Loss}.
\newblock {\em Econometric Theory}, 13(06):808--817.

\bibitem[\protect\citename{Cohn and Specia}2013]{Cohn2013}
Trevor Cohn and Lucia Specia.
\newblock 2013.
\newblock {Modelling Annotator Bias with Multi-task Gaussian Processes: An
  Application to Machine Translation Quality Estimation}.
\newblock In {\em Proceedings of ACL}, pages 32--42.

\bibitem[\protect\citename{Graham}2015]{Graham2015}
Yvette Graham.
\newblock 2015.
\newblock {Improving Evaluation of Machine Translation Quality Estimation}.
\newblock In {\em Proceedings of ACL}.

\bibitem[\protect\citename{Hensman \bgroup et al.\egroup }2013]{Hensman2013}
James Hensman, Nicol{\`{o}} Fusi, and Neil~D. Lawrence.
\newblock 2013.
\newblock {Gaussian Processes for Big Data}.
\newblock In {\em Proceedings of UAI}, pages 282--290.

\bibitem[\protect\citename{Hern{\'{a}}ndez-Lobato and
  Adams}2015]{Hernandez-Lobato2015}
Jos{\'{e}}~Miguel Hern{\'{a}}ndez-Lobato and Ryan~P. Adams.
\newblock 2015.
\newblock {Probabilistic Backpropagation for Scalable Learning of Bayesian
  Neural Networks}.
\newblock In {\em Proceedings of ICML}.

\bibitem[\protect\citename{Joshi \bgroup et al.\egroup }2010]{Joshi2010}
Mahesh Joshi, Dipanjan Das, Kevin Gimpel, and Noah~A. Smith.
\newblock 2010.
\newblock {Movie Reviews and Revenues: An Experiment in Text Regression}.
\newblock In {\em Proceedings of NAACL}.

\bibitem[\protect\citename{Koenker}2005]{Koenker2005}
Roger Koenker.
\newblock 2005.
\newblock {\em {Quantile Regression}}.
\newblock Cambridge University Press.

\bibitem[\protect\citename{Koponen \bgroup et al.\egroup }2012]{Koponen2012}
Maarit Koponen, Wilker Aziz, Luciana Ramos, and Lucia Specia.
\newblock 2012.
\newblock {Post-editing time as a measure of cognitive effort}.
\newblock In {\em Proceedings of WPTP}.

\bibitem[\protect\citename{L{\'{a}}zaro-Gredilla}2012]{Lazaro-Gredilla2012}
Miguel L{\'{a}}zaro-Gredilla.
\newblock 2012.
\newblock {Bayesian Warped Gaussian Processes}.
\newblock In {\em Proceedings of NIPS}, pages 1--9.

\bibitem[\protect\citename{Mihalcea and Strapparava}2012]{Mihalcea2012}
Rada Mihalcea and Carlo Strapparava.
\newblock 2012.
\newblock {Lyrics, Music, and Emotions}.
\newblock In {\em Proceedings of the Joint Conference on Empirical Methods in
  Natural Language Processing and Computational Natural Language Learning},
  pages 590--599.

\bibitem[\protect\citename{Newey and Powell}1987]{Newey1987}
Whitney~K. Newey and James~L. Powell.
\newblock 1987.
\newblock {Asymmetric Least Squares Estimation and Testing}.
\newblock {\em Econometrica}, 55(4).

\bibitem[\protect\citename{Nguyen and O'Connor}2015]{Nguyen2015}
Khanh Nguyen and Brendan O'Connor.
\newblock 2015.
\newblock {Posterior Calibration and Exploratory Analysis for Natural Language
  Processing Models}.
\newblock In {\em Proceedings of EMNLP}, number September, page~15.

\bibitem[\protect\citename{Qui{\~{n}}onero-Candela \bgroup et al.\egroup
  }2006]{Quinonero-Candela2006}
Joaquin Qui{\~{n}}onero-Candela, Carl~Edward Rasmussen, Fabian Sinz, Olivier
  Bousquet, and Bernhard Sch{\"{o}}lkopf.
\newblock 2006.
\newblock {Evaluating Predictive Uncertainty Challenge}.
\newblock {\em MLCW 2005, Lecture Notes in Computer Science}, 3944:1--27.

\bibitem[\protect\citename{Rasmussen and Williams}2006]{Rasmussen2006}
Carl~Edward Rasmussen and Christopher K.~I. Williams.
\newblock 2006.
\newblock {\em {Gaussian processes for machine learning}}, volume~1.
\newblock MIT Press Cambridge.

\bibitem[\protect\citename{Shah \bgroup et al.\egroup }2013]{Shah2013}
Kashif Shah, Trevor Cohn, and Lucia Specia.
\newblock 2013.
\newblock {An Investigation on the Effectiveness of Features for Translation
  Quality Estimation}.
\newblock In {\em Proceedings of MT Summit XIV}.

\bibitem[\protect\citename{Snelson \bgroup et al.\egroup }2004]{Snelson2004}
Edward Snelson, Carl~Edward Rasmussen, and Zoubin Ghahramani.
\newblock 2004.
\newblock {Warped Gaussian Processes}.
\newblock In {\em Proceedings of NIPS}.

\bibitem[\protect\citename{Snover \bgroup et al.\egroup }2006]{Snover2006}
Matthew Snover, Bonnie Dorr, Richard Schwartz, Linnea Micciulla, and John
  Makhoul.
\newblock 2006.
\newblock {A study of translation edit rate with targeted human annotation}.
\newblock In {\em Proceedings of AMTA}.

\bibitem[\protect\citename{Specia \bgroup et al.\egroup }2009]{Specia2009}
Lucia Specia, Nicola Cancedda, Marc Dymetman, Marco Turchi, and Nello
  Cristianini.
\newblock 2009.
\newblock {Estimating the sentence-level quality of machine translation
  systems}.
\newblock In {\em Proceedings of EAMT}, pages 28--35.

\bibitem[\protect\citename{Specia \bgroup et al.\egroup }2015]{Specia2015}
Lucia Specia, Gustavo~Henrique Paetzold, and Carolina Scarton.
\newblock 2015.
\newblock {Multi-level Translation Quality Prediction with QUEST++}.
\newblock In {\em Proceedings of ACL Demo Session}, pages 850--850.

\bibitem[\protect\citename{Specia}2011]{Specia2011}
Lucia Specia.
\newblock 2011.
\newblock {Exploiting Objective Annotations for Measuring Translation
  Post-editing Effort}.
\newblock In {\em Proceedings of EAMT}, pages 73--80.

\bibitem[\protect\citename{Strapparava and Mihalcea}2008]{Strapparava2008}
Carlo Strapparava and Rada Mihalcea.
\newblock 2008.
\newblock {Learning to identify emotions in text}.
\newblock In {\em Proceedings of the 2008 ACM Symposium on Applied Computing},
  pages 1556--1560.

\bibitem[\protect\citename{Zellner}1986]{Zellner1986}
Arnold Zellner.
\newblock 1986.
\newblock {Bayesian Estimation and Prediction Using Asymmetric Loss Functions}.
\newblock {\em Journal of the American Statistical Association},
  81(394):446--451.

\end{thebibliography}
\bibliographystyle{acl2016}

\end{document}